\documentclass{article}

\usepackage{spconf,amsmath,graphicx}
\usepackage[hang,flushmargin]{footmisc} 
\usepackage{amsmath}
\usepackage{amssymb}
\usepackage{tabularx}
\usepackage{multirow}
\usepackage{caption} 
\usepackage{booktabs}
\usepackage{adjustbox}
\usepackage{float}
\usepackage{array}
\usepackage{makecell}  
\usepackage[colorlinks=true, linkcolor=blue, citecolor=blue, urlcolor=blue]{hyperref}
\hypersetup{draft}            
\usepackage{cite} 

\usepackage{enumitem}
\setlist{nosep, leftmargin=14pt}

\usepackage{mwe} 

\vspace{-0.5cm}
\title{LUMOS: Universal Semi-Supervised OCT Retinal Layer Segmentation with Hierarchical Reliable Mutual Learning}
\name{Yizhou Fang$^{1 \dagger}$ \qquad Jian Zhong$^{1,2\dagger}$ \qquad Li Lin$^{1\dagger}$  \qquad Xiaoying Tang$^{1,3 \star}$
\thanks{%
\parindent=0pt
\textsuperscript{†} These authors contributed equally.\\
\textsuperscript{*} Corresponding author: Dr.\ Xiaoying Tang (\href{mailto:tangxy@sustech.edu.cn}{\textcolor{blue}{tangxy@sustech.edu.cn}}).
}
}

\address{
\textsuperscript{1}Department of Electronic and Electrical Engineering,
Southern University of Science and Technology,\\ Shenzhen, China \\
\textsuperscript{2}Shenzhen Loop Area Institute, Shenzhen, China\\
\textsuperscript{3}Jiaxing Research Institute, Southern University of Science and Technology, Jiaxing, China
}

\begin{document}
%
\maketitle
\vspace{-0.8cm}
\begin{abstract}
Optical Coherence Tomography (OCT) layer segmentation faces challenges due to annotation scarcity and heterogeneous label granularities across datasets. While semi-supervised learning helps alleviate label scarcity, existing methods typically assume a fixed granularity, failing to fully exploit cross-granularity supervision. This paper presents LUMOS, a semi-supervised universal OCT retinal layer segmentation framework based on a Dual-Decoder Network with a Hierarchical Prompting Strategy (DDN-HPS) and Reliable Progressive Multi-granularity Learning (RPML). DDN-HPS combines a dual-branch architecture with a multi-granularity prompting strategy to effectively suppress pseudo-label noise propagation. Meanwhile, RPML introduces region-level reliability weighing and a progressive training approach that guides the model from easier to more difficult tasks, ensuring the reliable selection of cross-granularity consistency targets, thereby achieving stable cross-granularity alignment. Experiments on six OCT datasets demonstrate that LUMOS largely outperforms existing methods and exhibits exceptional cross-domain and cross-granularity generalization capability.

\end{abstract}

\begin{keywords}
OCT, universal segmentation, multi-granularity,  semi-supervised learning, dual-branch mutual learning
\end{keywords}
\vspace{-0.3cm}
\section{Introduction}
\vspace{-0.2cm}
Optical Coherence Tomography (OCT) provides high-resolution cross-sectional images that are crucial for diagnosing ophthalmic diseases like diabetic retinopathy and glaucoma~\cite{araki2022optical}. Accurate retinal layer segmentation~\cite{tan2023retinal} is fundamental to clinical diagnosis but faces significant challenges: manual segmentation is time-consuming with high inter-observer variability. In recent years, deep learning methods have achieved remarkable success in OCT retinal layer segmentation, enabling precise delineation. However, deep learning based segmentation methods heavily rely on large-scale high-quality annotations. Expert OCT layer annotation is not only costly, but also exhibits varying annotation granularities across datasets due to different research objectives and clinical requirements, posing challenges for unified modeling. Meanwhile, abundant unlabeled OCT data accumulated from clinical practice remains underutilized~\cite{rozhyna2024exploring}.

Semi-supervised learning mitigates label scarcity by jointly exploiting labeled and unlabeled data. Recent advances include: DiffRect~\cite{liu2024diffrect} corrects pseudo-label distributions via latent diffusion; CGS~\cite{wang2025balancing} adopts a specialist–generalist collaboration for multi-target segmentation; and ABD~\cite{chi2024adaptive} strengthens consistency with adaptive bidirectional displacement. Nevertheless, these methods typically assume a fixed annotation granularity. When datasets differ in granularity, they often require separate models, hindering efficient label utilization. It is worth noting that OCT datasets commonly exhibit heterogeneous granularities—some annotate all retinal layers, whereas others mark only key pathological boundaries—making single-granularity pipelines ill-suited to exploit cross-granularity supervision.

To address these challenges, we propose LUMOS—a semi-supervised universal OCT retinal layer segmentation framework based on a Dual-Decoder Network with a Hierarchical Prompting Strategy (DDN-HPS) and Reliable Progressive Multi-granularity Learning (RPML). This work's main contributions are three-fold: (1) To the best of our knowledge, we introduce the first semi-supervised universal framework for OCT retinal layer segmentation. We employ dual-branch mutual learning to effectively utilize multi-source data, thereby establishing a unified approach for handling domain shifts and granularity variations. (2) We propose DDN-HPS to suppress pseudo-label noise through dual-branch mutual supervision with multi-granularity prompting, and develop RPML to ensure prediction reliability through region-level reliability weighing and a progressive training approach. (3) Extensive experiments on six OCT datasets across multiple granularities demonstrate the superiority of LUMOS over state-of-the-art semi-supervised methods, particularly under limited annotation settings.
\vspace{-3ex}
\begin{figure*}[t]
\vspace{-0.8cm}
\centering
\includegraphics[width=1.0\linewidth]{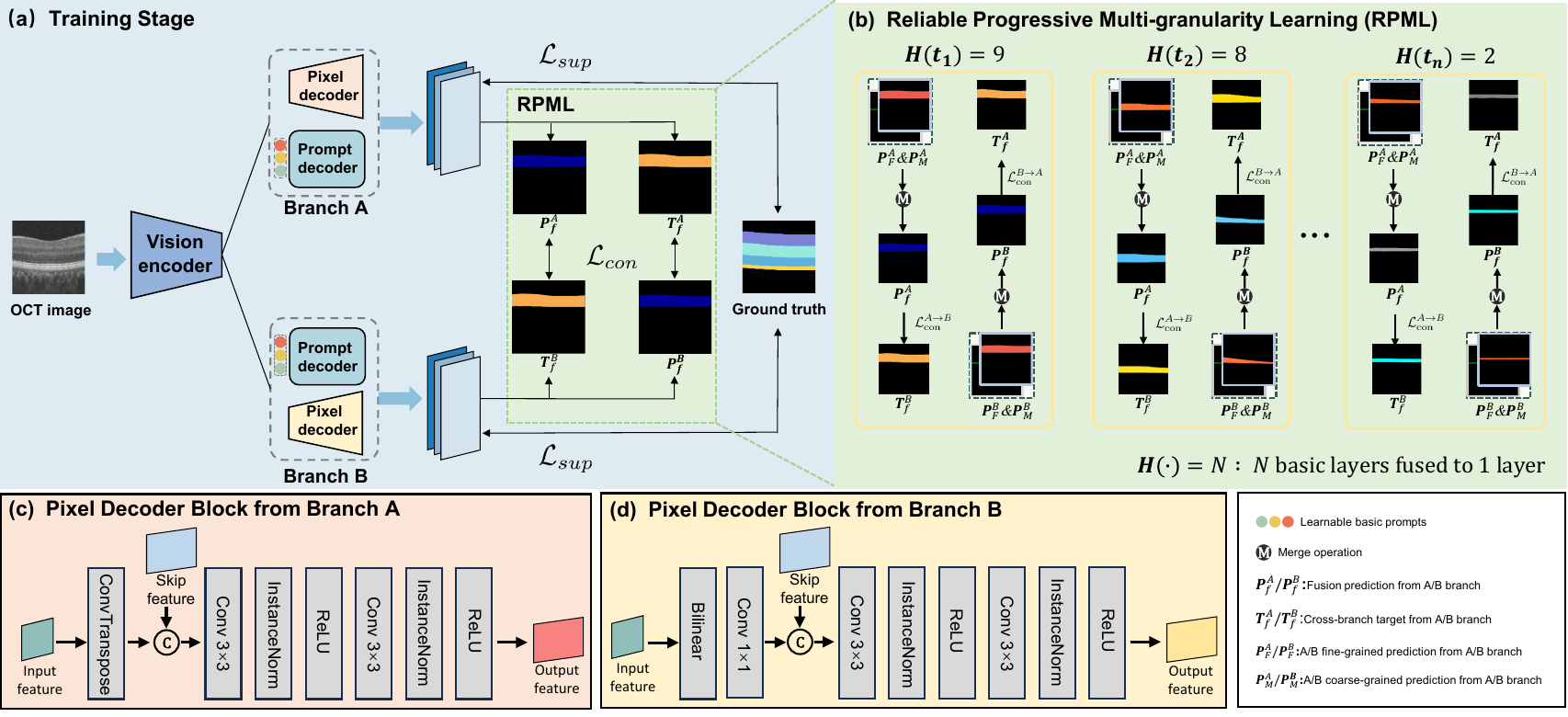} 
\caption{Overview of the proposed LUMOS framework and its key components.}
\label{fig:1}

\end{figure*}

\section{Method}
\subsection{LUMOS Paradigm Overview}
Conventional semi-supervised methods typically address a single annotation granularity, training $N$ separate models for $N$ tasks involving different annotation granularities. This hinders label reuse across granularities and inflates computation and storage, ultimately limiting performance and generalization. LUMOS unifies multi-granularity labeled data and unlabeled data within a single training paradigm, enabling cross-granularity segmentation via dual-branch mutual learning and cross-granularity consistency constraining.

Formally, let $\{D_i\}_{i=1}^{N}$ denote datasets with heterogeneous retinal-layer granularities, where each $D_i=\{(x,y)\}$ contains OCT images $x$ (width $W$, height $H$) and corresponding labels $y$. We define the granularity of $D_i$ as $\mathcal{G}(D_i)$, i.e., the number of annotated OCT layers.

The LUMOS framework (Fig.~\textcolor{blue}{\ref{fig:1}}) employs a shared encoder with dual decoders and two key modules: DDN-HPS suppresses pseudo-label noise via dual-branch mutual supervision, while RPML enforces cross-granularity consistency through dynamic task scheduling. The objective combines supervised (Dice + BCE) and consistency terms:
\begin{equation}
\mathcal{L}_{\mathbf{total}} = \mathcal{L}_{\mathbf{sup}} + \mathcal{L}_{\mathbf{con}}.
\end{equation}

\subsection{Dual-Decoder Network with a Hierarchical Prompting Strategy}

We employ a dual-decoder architecture $\mathbf{De}_i$, $i \in \{A, B\}$, with a shared hybrid CNN--Transformer encoder $E(\cdot)$. From an input OCT image $\mathbf{I} \in \mathbb{R}^{H \times W}$, the encoder extracts features $\mathbf{F} = E(\mathbf{I}) \in \mathbb{R}^{C \times H' \times W'}$. Each decoder comprises: (i) a pixel decoder that mirrors the encoder and uses skip connections to upsample multi-scale features into the final retinal feature map $\mathbf{F}_p \in \mathbb{R}^{C_{out} \times H \times W}$, where $\mathbf{De}_A$ employs transposed convolution for adaptable learnable upsampling, while $\mathbf{De}_B$ utilizes fixed, parameter-free bilinear interpolation as a stable auxiliary branch; and (ii) a prompt decoder that processes multi-stage outputs $\{\mathbf{F}_s^k\}_{k=1}^5$ for prompt learning.

The prompt decoder is built from Transformer decoder layers with masked attention and self-attention~\cite{cheng2022masked}. Following UniOCTSeg’s multi-granularity prompting~\cite{zhong2025unioctseg}, we instantiate nine learnable basic prompts $\bigcup_{l=1}^{9} \mathbf{P}_b^l \in \mathbb{R}^{\frac{W}{32} \times \frac{H}{32}}$ corresponding to the finest-grained OCT layer segmentation~\cite{he2018retinal}. These prompts are updated by the prompt decoder and combined to represent layers of arbitrary granularity, thus unifying heterogeneous annotation schemes within a single framework.

Finally, inter-branch mutual supervision between $\mathbf{De}_{A}$ and $\mathbf{De}_{B}$ suppresses pseudo-label noise and alleviates the overfitting tendency commonly observed in single-branch frameworks. Inference relies solely on $\mathbf{De}_A$.
\vspace{-0.2cm}
\subsection{Reliable Progressive Multi-granularity Learning}

To ensure reliable predictions across granularities, we design RPML, which establishes cross-granularity consistency through progressive multi-level representation learning.

First, RPML introduces a mutual learning loss. Each branch leverages the pseudo-labels generated by its counterpart, enforcing bidirectional consistency:
\begin{equation}
\mathcal{L}^{(A \rightarrow B)} = \mathbf{BCE}(\mathbf{P_s}^{(A)}, \mathbf{T}^{(B)}),
\end{equation}
where \( \mathbf{P_s} \) denotes the student predictions and \( \mathbf{T} \) represents pseudo-labels provided by the teacher branch. The overall mutual learning loss is then defined as:
\begin{equation}
\mathcal{L}_{\mathbf{ML}} = \mathcal{L}^{(A \rightarrow B)} + \mathcal{L}^{(B \rightarrow A)}.
\end{equation}

Pseudo-labels are inherently susceptible to noise. To mitigate this, RPML introduces
Region-level Reliability Weighing (RLRW), defined as:
\begin{equation}
\mathbf{W}_{\text{region}} = \mathbf{M} \odot \left( \mathbf{1} - (\mathbf{S}_{\text{sim}} - \mathbf{P}_{\text{t}})^2 \right),
\end{equation}
where $\mathbf{S}_{\mathbf{sim}}$ denotes feature–prototype similarity, $\mathbf{P_t}$ represents teacher predictions, and $\mathbf{M}$ highlights pixels with higher uncertainty in $\mathbf{P_s}$ than $\mathbf{P_t}$, whereby RLRW suppresses unreliable regions and enhances robustness.

Building upon RLRW, RPML performs Multi-level Prediction Fusion (MPF). Specifically, RPML identifies a coarse-grained prediction $\mathbf{P_M}$ and merges it with a fine-grained prediction $\mathbf{P_F}$ to construct a fused prediction $\mathbf{P_f}$ to align with the target $\mathbf{T_f}$:
\begin{equation}
\mathbf{P_f} = \mathbf{Merge}(\mathbf{P_M}, \mathbf{P_F}),
\end{equation}
thus ensuring cross-granularity structural consistency.  

By combining the above components, the unidirectional consistency loss is formulated as:
\begin{equation}
\mathcal{L}_{\mathbf{con}}^{(A \rightarrow B)} = 
\frac{\sum_{h,w} \mathbf{W}_{\mathbf{region},h,w} \cdot \mathcal{L}^{(A \rightarrow B)}}{\sum_{h,w} \mathbf{M}_{h,w} + \epsilon}.
\end{equation}
where $h, w$ denote spatial coordinates. To achieve symmetric consistency across branches and granularities, the final RPML loss is defined as:
\begin{equation}
\mathcal{L}_{\mathbf{con}} = \mathcal{L}_{\mathbf{con}}^{(A \rightarrow B)} + \mathcal{L}_{\mathbf{con}}^{(B \rightarrow A)}.
\end{equation}

This loss is computed symmetrically in both directions, achieving consistency across branches and granularities.  

Finally, as training progresses, $\mathbf{T_f}$ is progressively adjusted, correspondingly updating task complexity. Early stages focus on simpler coarse-grained tasks to ensure structural stability, while later stages tackle more complex fine-grained tasks to learn hierarchical representations.

\section{EXPERIMENTS}
\label{sec:pagestyle}
\subsection{Dataset and Preprocessing} 
We assemble multiple multi-granularity OCT layer segmentation datasets acquired across different devices as the internal sets: HC-MS~\cite{he2018retinal}, GCN~\cite{li2021multi}, and OCTA-500~\cite{li2024octa}, containing 3,430, 482, and 9,416 samples with 8, 8, and 5 granularity levels, respectively. HC-MS~\cite{he2018retinal} and GCN~\cite{li2021multi} are split into training, validation, and test sets, and unlabeled data from OCTA-500~\cite{li2024octa} are added to the training set at an approximate 2:8 labeled-to-unlabeled ratio for semi-supervised learning. Four additional public datasets—HEG~\cite{tian2015real}, Goals~\cite{fang2022dataset}, AMD~\cite{chiu2012validated}, and OIMHS~\cite{ye2023oimhs}—comprising 100, 100, 220, and 2,805 samples with 7, 3, 2, and 1 granularity levels, respectively, are used as the external test sets (Table~\textcolor{blue}{\ref{tab:dataset}}). All images are resized to 512×512 and normalized. During training, data augmentations including horizontal flipping, random Gaussian noise, and random brightness and contrast adjustments are applied.

\begin{table}[htbp]
\vspace{-0.2cm}
\centering
\caption{Dataset statistics.}
\label{tab:dataset}
\renewcommand{\arraystretch}{0.8}
\small
\adjustbox{width=\columnwidth}{%
\begin{tabular}{cccccc} 
\toprule
\multicolumn{2}{c}{Dataset} & Type & Device & Scan Num & \begin{tabular}[c]{@{}c@{}}Annotation\\ Granularity\end{tabular} \\ 
\midrule
\multirow{3}{*}{\begin{tabular}[c]{@{}c@{}}Internal\\ set\end{tabular}} & HC-MS~\cite{he2018retinal} & \multirow{2}{*}{Labeled} & Heidelberg & 3430 & 8 \\
& GCN~\cite{li2021multi} & & Topcon & 482 & 8 \\
& OCTA-500~\cite{li2024octa} & Unlabeled & Optovue & 9416 & 5 \\ 
\midrule
\multirow{4}{*}{\begin{tabular}[c]{@{}c@{}}External\\ set\end{tabular}} & HEG~\cite{tian2015real} & \multirow{4}{*}{Labeled} & Heidelberg & 100 & 7 \\
& Goals~\cite{fang2022dataset} & & Topcon & 100 & 3 \\
& AMD~\cite{chiu2012validated} & & Bioptigen & 220 & 2 \\
& OIMHS~\cite{ye2023oimhs} & & Topcon & 2805 & 1 \\ 
\bottomrule
\end{tabular}%
}
\vspace{-0.5cm}
\end{table}

\subsection{Implementation Details} All experiments are implemented using PyTorch on an A6000 GPU. During training, we employ the Adam optimizer with a learning rate of 3e-4, a batch size of 8, and a total of 80,000 iterations. The learning rate is annealed at each iteration using a polynomial decay schedule with a power of 0.9.

\begin{figure*}[htb]
\vspace{-1cm}
\centering
\includegraphics[width=0.9\textwidth]{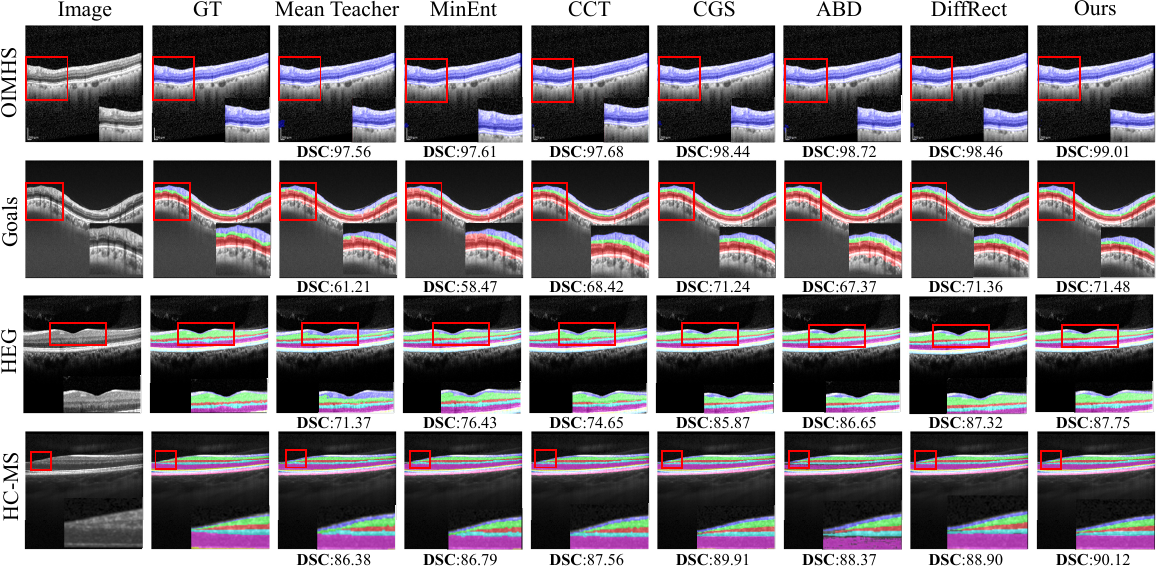}  
\caption{Visualization comparison results on the OIMHS, Goals, HEG and HC-MS datasets.}
\label{fig:2}
\end{figure*}

\begin{table*}[!htbp]
\centering
\caption{Comparison results on the internal and external datasets.}
\label{tab:comparison}
\renewcommand{\arraystretch}{1.5}
\adjustbox{width=\textwidth}{%
\begin{tabular}{lcccccccccccccc}
\toprule
\multirow{3}{*}{\textbf{Method}} & \multicolumn{4}{c}{\textbf{Internal Datasets}} & \multicolumn{8}{c}{\textbf{External Datasets}} & \multicolumn{2}{c}{}\\
\cmidrule(lr){2-5} \cmidrule(lr){6-13}
 & \multicolumn{2}{c}{HC-MS} & \multicolumn{2}{c}{GCN} & \multicolumn{2}{c}{HEG} & \multicolumn{2}{c}{Goals} & \multicolumn{2}{c}{AMD} & \multicolumn{2}{c}{OIMHS} & \multicolumn{2}{c}{Avg.} \\
 \cmidrule(lr){2-5} \cmidrule(lr){6-13}\cmidrule(lr){14-15}
 & DSC ↑ & HD95 ↓ & DSC ↑ & HD95 ↓ & DSC ↑ & HD95 ↓ & DSC ↑ & HD95 ↓ & DSC ↑ & HD95 ↓ & DSC ↑ & HD95 ↓ & DSC ↑ & HD95 ↓ \\
\midrule
Mean Teacher(2017)~\cite{tarvainen2017mean} & 86.59 ± 7.87 & 11.70 ± 4.08 & 71.55 ± 12.57 & 11.61 ± 3.88 & 80.24 ± 9.94 & 6.19 ± 6.57 & 59.52 ± 31.04 & 45.13 ± 36.77 & 55.78 ± 14.07 & 46.28 ± 37.25 & 91.76 ± 12.08 & 13.87 ± 8.41 & 74.24 ± 14.56 & 22.46 ± 18.18 \\
CCT (2020)~\cite{ouali2020semi} & 86.65 ± 4.06 & 3.61 ± 0.84 & 76.71 ± 1.00 & 6.35 ± 2.04 & 78.09 ± 8.44 & 6.53 ± 5.59 & 66.02 ± 25.44 & 38.42 ± 36.23 & 51.05 ± 36.19 & 36.98 ± 27.78 & 93.19 ± 12.77 & 11.86 ± 13.90 & 75.29 ± 15.05 & 17.29 ± 16.04 \\
MinEnt (2019)~\cite{vu2019advent} & 87.42 ± 3.85 & 4.23 ± 0.74 & 76.55 ± 8.99 & 6.46 ± 3.62 & 82.48 ± 8.40 & 4.88 ± 3.30 & 66.78 ± 23.58 & 25.61 ± 22.47 & 62.30 ± 26.30 & 45.22 ± 20.20 & 95.61 ± 7.03 & 7.54 ± 11.04 & 78.52 ± 12.58 & 15.66 ± 16.55 \\
DiffRect (2024)~\cite{liu2024diffrect} & 88.30 ± 3.81 & 2.90 ± 0.39 & 77.96 ± 8.87 & 5.26 ± 3.41 & 85.60 ± 3.39 & 6.62 ± 6.54 & 72.29 ± 11.12 & 9.77 ± 7.54 & 77.97 ± 12.33 & 8.28 ± 12.55 & 95.94 ± 3.35 & 3.87 ± 8.41 & 83.01 ± 8.58 & 6.12 ± 2.62 \\
ABD (2024)~\cite{chi2024adaptive} & 88.19 ± 5.34 & 2.39 ± 1.65 & 79.45 ± 12.65 & 4.79 ± 5.05 & 85.45 ± 5.63 & 4.33 ± 3.28 & 72.79 ± 12.73 & 12.56 ± 9.67 & 73.98 ± 18.41 & 4.33 ± 3.28 & 97.37 ± 1.15 & 2.76 ± 9.18 & 82.87 ± 9.35 & 5.19 ± 3.73 \\
CGS (2024)~\cite{wang2025balancing} & 88.90 ± 3.42 & 2.30 ± 0.16 & 80.98 ± 6.79 & 4.62 ± 2.01 & 84.56 ± 4.69 & 3.08 ± 0.27 & 68.28 ± 10.69 & 16.39 ± 12.79 & 77.33 ± 10.33 & 13.85 ± 7.81 & 96.14 ± 4.12 & 5.04 ± 7.53 & 82.70 ± 9.62 & 7.55 ± 6.00 \\
\textbf{Ours-LUMOS} & \textbf{90.84 ± 4.05} & \textbf{0.91 ± 0.57} & \textbf{81.72 ± 10.30} & \textbf{2.55 ± 2.69} & \textbf{87.81 ± 3.63} & \textbf{1.27 ± 0.48} & \textbf{73.69 ± 14.59} & \textbf{6.74 ± 9.24} & \textbf{84.30 ± 10.28} & \textbf{3.29 ± 4.42} & \textbf{97.96 ± 1.45} & \textbf{0.64 ± 6.79} & \textbf{86.05 ± 8.28} & \textbf{2.57 ± 2.28} \\
\bottomrule
\end{tabular}%
}
\end{table*}
\vspace{-0.5cm}
\subsection{Comparisons with State-of-the-art}

We benchmark LUMOS against several representative semi-supervised frameworks, including classical methods such as Mean Teacher~\cite{tarvainen2017mean}, CCT~\cite{ouali2020semi}, and MinEnt~\cite{vu2019advent}, as well as recent approaches such as DiffRect~\cite{liu2024diffrect}, ABD~\cite{chi2024adaptive}, and CGS~\cite{wang2025balancing}. All methods are evaluated using the Dice Similarity Coefficient (DSC) and the 95\% Hausdorff Distance (HD95).

As shown in Table~\textcolor{blue}{\ref{tab:comparison}}, LUMOS demonstrates superior performance on both internal test sets. On the HC-MS dataset~\cite{he2018retinal}, our method achieves an average DSC of 90.84\% with an average HD95 of 0.91 pixels, exceeding the second-best approach CGS~\cite{wang2025balancing} by 1.94\% in DSC and reducing HD95 by 1.39 pixels. Similarly, on the GCN dataset~\cite{li2021multi}, LUMOS obtains an average DSC of 81.72\% and an average HD95 of 2.55 pixels, outperforming CGS by 0.74\% in DSC while decreasing HD95 by 2.07 pixels. The consistent performance gains across these two internal datasets with distinct structural granularities confirm LUMOS's strong capability in cross-granularity segmentation tasks.

To further evaluate cross-granularity adaptability and generalization capability, all methods are evaluated on the external test sets featuring granularity configurations absent from the training data. For fair comparison, HC-MS~\cite{he2018retinal} is reconstructed by merging selected retinal layers to match the granularity of each external dataset, and all methods are retrained accordingly. As summarized in Table~\textcolor{blue}{\ref{tab:comparison}}, LUMOS surpasses the second-best method by 2.21\%, 0.90\%, 6.33\%, and 0.59\% in DSC on HEG~\cite{tian2015real}, Goals~\cite{fang2022dataset}, AMD~\cite{chiu2012validated}, and OIMHS~\cite{ye2023oimhs}, respectively, while reducing HD95 by 1.81, 3.03, 1.04, and 2.12 pixels. Overall, LUMOS achieves superior performance with an average DSC of 86.05\% and an average HD95 of 2.57 pixels across all internal and external datasets, demonstrating consistent improvements of 3.04\% in DSC and 2.62 pixels in HD95 compared to the second-best method, verifying its robust generalization to heterogeneous external data distributions. Representative visualizations are provided in Fig.~\textcolor{blue}{\ref{fig:2}}.

\begin{table}[t]

\centering
\caption{Ablation results of the dual-branch design.}
\label{tab:ablation}
\scriptsize
\resizebox{1.0\linewidth}{!}{%
\begin{tabular}{ccccccc}
\toprule
\multirow{2}{*}{\textbf{Setting}} & 
\multicolumn{2}{c}{\textbf{Internal Datasets}} & \multicolumn{2}{c}{\textbf{External Datasets}} & \multicolumn{2}{c}{\textbf{Avg.}} \\
& DSC ↑& HD95 ↓& DSC ↑& HD95 ↓& DSC ↑& HD95 ↓\\ 
\midrule
A1 & 85.13 ± 2.08 & 3.97 ± 9.77 & 82.92 ± 8.25 & 22.34 ± 23.04 & 84.03 ± 5.17 & 13.16 ± 16.41 \\
A2 & 85.15 ± 7.38 & 3.98 ± 7.86 & 83.51 ± 7.81 & 16.58 ± 22.38 & 84.33 ± 7.60 & 10.28 ± 15.12 \\
\bottomrule
\end{tabular}%

}

\vspace{2pt}
\raggedright\scriptsize
A1: conventional teacher–student strategy with $\mathcal{L}^{(s\rightarrow t)}$ \\
A2: our network with $\mathcal{L}_{\mathrm{ML}}$

\end{table}

\subsection{Ablation Studies}

We conduct ablation studies to validate the effectiveness of each component in LUMOS.

\noindent\textbf{(1) Ablation of the dual-branch design.}
We compare our dual-branch mutual learning approach with the traditional teacher–student paradigm under the following settings: A1 (conventional teacher–student strategy with $\mathcal{L}^{(s\rightarrow t)}$) and A2 (our network with $\mathcal{L}_{\mathrm{ML}}$). As shown in Table~\textcolor{blue}{\ref{tab:ablation}}, A2 achieves an 0.30\% increase in average DSC and a 2.88-pixel reduction in HD95, demonstrating the effectiveness of the dual-branch mutual learning mechanism.

\noindent\textbf{(2) Ablation of RPML's key components.}
RPML comprises two key components: RLRW and MPF. As demonstrated in Table~\textcolor{blue}{\ref{tab:ablation_rpml}}, adding RLRW to the dual-branch framework improves the average DSC by 1.51\% and reduces the average HD95 by 1.00 pixel, showing its effectiveness in suppressing noisy pseudo-labels. Further incorporating MPF yields an additional 0.27\% improvement in the average DSC, and the HD95 is substantially reduced by 6.92 pixels.

\begin{table}[]
\centering
\caption{Ablation results of RPML's key components.}
\label{tab:ablation_rpml}
\setlength{\tabcolsep}{3pt}
\renewcommand{\arraystretch}{0.9}
\footnotesize
\resizebox{1.0\linewidth}{!}{%
\begin{tabular}{cccccccc}
\toprule
\multicolumn{2}{c}{\textbf{Components}} & \multicolumn{2}{c}{\textbf{Internal Datasets}} & \multicolumn{2}{c}{\textbf{External Datasets}} & \multicolumn{2}{c}{\textbf{Avg.}} \\
RLRW & MPF & DSC $\uparrow$ & HD95 $\downarrow$ & DSC $\uparrow$ & HD95 $\downarrow$ & DSC $\uparrow$ & HD95 $\downarrow$ \\
\midrule
$-$ & $-$ & 85.15 $\pm$ 7.38 & 3.98 $\pm$ 7.86 & 83.51 $\pm$ 7.81 & 16.58 $\pm$ 22.38 & 84.33 $\pm$ 7.60 & 10.28 $\pm$ 15.12 \\
$\checkmark$ & $-$ & 85.94 $\pm$ 2.31 & 1.98 $\pm$ 2.02 & 85.73 $\pm$ 9.14 & 11.52 $\pm$ 19.38 & 85.84 $\pm$ 5.73 & 9.28 $\pm$ 12.20 \\
$\checkmark$ & $\checkmark$ & 86.28 $\pm$ 7.83 & 1.73 $\pm$ 1.94 & 85.94 $\pm$ 9.14 & 2.99 $\pm$ 6.15 & 86.11 $\pm$ 8.49 & 2.36 $\pm$ 4.05 \\
\bottomrule
\end{tabular}%
}
\end{table}

\section{CONCLUSION}
\vspace{-0.1cm}
\label{sec:typestyle}

This study proposes LUMOS, a semi-supervised universal framework for multi-granularity OCT retinal layer segmentation. Experiments on six datasets demonstrate that LUMOS and its components significantly outperform existing methods in cross-domain and cross-granularity adaptation. However, limitations persist regarding handling severe pathological deformations. Future work will focus on addressing this challenge to further enhance the framework's robustness and clinical usability.

\section{Acknowledgments}

\begin{sloppypar}
This study was supported by the National Key Research and Development Program of China (2023YFC2415400); the National Natural Science Foundation of China (T2422012); the Guangdong Basic and Applied Basic Research (2024B1515020088); the High Level of Special Funds (G030230001, G03034K003); the Guangdong Key Research and Development Program (2025B1111080001); the SUSTech Fang Keng Faculty Award.
\end{sloppypar}

\section{Compliance with ethical standards}
This research study was conducted retrospectively using publicly available and anonymized 
datasets (OCTA-500, HC-MS, GCN,  AMD, HEG, GOALS, OIMHS). Ethical approval was not 
required as confirmed by the open data usage terms governing these datasets.

\bibliographystyle{IEEEbib}   
\bibliography{strings,refs}   

\end{document}